\pdfoutput=1

\documentclass[11pt]{article}

\usepackage[preprint]{acl}
\usepackage{times}
\usepackage{latexsym}
\usepackage{booktabs}
\usepackage[T1]{fontenc}
\usepackage[utf8]{inputenc}
\usepackage{microtype}
\usepackage{inconsolata}
\usepackage{graphicx}
\usepackage{amsmath}
\usepackage{tcolorbox}
\usepackage{hyperref}
\usepackage[capitalise]{cleveref}
\usepackage{subcaption}
\usepackage{multirow}
\usepackage{pgfplots}
\pgfplotsset{compat=1.18}

\title{SECQUE: A Benchmark for Evaluating Real-World Financial Analysis Capabilities}

\author{
    \makebox[.2\linewidth]{Noga Ben Yoash
    \thanks{Corresponding author: nogabenyoash@microsoft.com}} \hspace{0.5cm} 
    \makebox[.2\linewidth]{Meni Brief}
    \AND
    \makebox[.2\linewidth]{Oded Ovadia} \hspace{0.5cm}
    \makebox[.2\linewidth]{Gil Shenderovitz} \hspace{0.5cm}
    \makebox[.2\linewidth]{Moshik Mishaeli} \hspace{0.5cm}
    \AND
    \makebox[.2\linewidth]{Rachel Lemberg} \hspace{0.5cm}
    \makebox[.2\linewidth]{Eitam Sheetrit} 
    \\\\
    \makebox[\linewidth]{Microsoft Industry AI} 
}

\begin{document}

\maketitle

\begin{abstract}
We introduce SECQUE, a comprehensive benchmark for evaluating large language models (LLMs) in financial analysis tasks. SECQUE comprises 565 expert-written questions covering SEC filings analysis across four key categories: comparison analysis, ratio calculation, risk assessment, and financial insight generation. To assess model performance, we develop SECQUE-Judge, an evaluation mechanism leveraging multiple LLM-based judges, which demonstrates strong alignment with human evaluations. Additionally, we provide an extensive analysis of various models' performance on our benchmark. By making SECQUE publicly available\footnote{\url{https://huggingface.co/datasets/nogabenyoash/SecQue}}, we aim to facilitate further research and advancements in financial AI.
\end{abstract}

\section{Introduction}
Recent advances in large language models (LLMs) have demonstrated their potential across diverse domains, including law~\cite{Huang2023}, medicine~\cite{Singhal2023, Wu2024}, and finance~\cite{Cheng2023, Wu2023}. However, as these models are increasingly adopted for specialized applications, the need for domain-specific evaluation has become more pressing. While general-purpose benchmarks assess a wide range of capabilities, they often fail to capture the nuances and challenges inherent in domain-specific tasks~\cite{yang2024harnessing}.

While domain-specific evaluation is challenging across many fields, the financial domain presents unique challenges in assessing LLM capabilities. Financial analysts routinely analyze complex datasets, extract meaningful insights from textual and numerical data, and answer high-stakes questions about companies, industries, and market trends. These tasks require models to excel in financial reasoning, numerical computation, and the synthesis of information from lengthy, multi-format documents. Yet, many existing benchmarks for financial LLMs often focus on isolated downstream tasks, such as sentiment analysis or named entity recognition, and do not adequately reflect the breadth of questions analysts face in real-world scenarios~\cite{xie2024pixiu, brief2024mixing, Islam2023}.

To address this gap, we introduce SECQUE, a benchmark specifically designed to evaluate LLMs on the types of questions financial analysts pose while analyzing SEC\footnote{SEC is the common name for the U.S. Securities and Exchange Commission} filings. SECQUE includes questions spanning four key categories: Comparison and Trend Analysis, Ratio Analysis, Risk Factors, and Analyst Insights, thus representing essential components of financial analysis. For each question, we present a ground truth answer and variations of the supporting data from the SEC filings, representing different textual pre-processing methods. The benchmark consists of 565 questions curated to challenge models' abilities to comprehend, reason, and synthesize information within the context of corporate filings. 

Our benchmark offers several key advantages. First, SECQUE is designed to reflect real-world financial tasks, moving beyond basic text processing to assess reasoning over long unstructured data. Second, it emphasizes long-context questions, requiring models to extract relevant information from complex and detailed inputs, such as financial tables with varied structures. Third, SECQUE addresses limitations identified in FinanceBench~\cite{Islam2023} by introducing cross-company comparisons and high-difficulty questions.

Additionally, following~\cite{Zheng2023}, LLM judges have become a central component of open-ended question evaluation, and SECQUE significantly relies on the ability to use LLMs for evaluation accordingly. The questions in SECQUE are of high complexity and therefore present difficulty for LLM judging. To address this difficulty, we present SECQUE-judge that, following~\cite{gu2024survey}, leverages multiple LLM judges evaluations. We perform a thorough investigation of SECQUE-judge and demonstrate its alignment with human evaluation. Using our validated SECQUE-judge, we have performed a thorough analysis of SECQUE. Finally, we conduct an ablation study to examine how different configurations, such as prompt choice and temperature, affect the results.

\begin{table}[t]
\centering
\setlength{\tabcolsep}{6pt}
\caption{Summary Statistics of the SEC filings used in SECQUE.}
\label{tab:summary_stats}
\begin{tabular}{p{5cm}c}
\toprule
\textbf{Statistic}                  & \textbf{Value} \\ 
\midrule
Unique Accessions                   & 45             \\ 
Unique Companies                    & 29             \\ 
Unique Filing Years                 & 4              \\ 
Companies with Multiple Filings     & 12              \\ 
Earliest Filing Date                & 7/25/2018     \\ 
Latest Filing Date                  & 8/8/2024       \\ 
\bottomrule
\end{tabular}
\end{table}

\section{SECQUE Benchmark}\label{sec:secque_benchmark}

The SECQUE benchmark was developed as a tool to evaluate the performance of large language models (LLMs) specializing in the financial domain in real-world financial scenarios. Our evaluation focuses on key use cases where LLMs could significantly impact the work of financial professionals in general, and financial analysts in particular. Financial analysts rely on diverse documents in their work, and we focused on the primary publicly available financial reports\footnote{\url{https://sec.gov/edgar/search}}: 10-K and 10-Q SEC filings. A 10-K is a company's annual financial report filed with the SEC, while a 10-Q is a quarterly update on its financial performance. These documents include textual and tabular data about publicly traded companies, covering sections such as risk factors, income statements, balance sheets, and cash flow statements. 

\noindent\textbf{Benchmark Creation:}
The SECQUE benchmark was created by three subject matter experts (SMEs) specializing in financial analysis.
To ensure high standards, all questions and answers were iteratively refined and reviewed both by the SMEs and by two additional financial experts with expertise in LLM systems.

\noindent\textbf{Benchmark Composition:} The benchmark consists of 565 open-ended questions representing real-world financial analysts' questions in terms of complexity, jargon, and type. Each entry in the benchmark includes a question, supporting data (also referred to as context), and a ground truth answer. Additionally, references to the supporting data (e.g., metadata specifying accession numbers, page numbers, and relevant sections from the filings that indicate the source of the context) and a question type label are provided. 

Following is an example data point from SECQUE benchmark (for full context see ~\cref{fig:example_question_ratio}).\\

\begin{figure}[!htb]
\centering
\resizebox{0.45\textwidth}{!}{
\begin{tcolorbox}[colframe=black!40, colback=black!5!white, sharp corners=southwest, boxrule=0.6mm, toptitle=1mm, bottomtitle=1mm]
\small
\vspace{0mm}
\textbf{Input:}
\begin{itemize}
    \setlength{\itemsep}{0pt}
    \item \textbf{Question:} 
    \texttt{How has NVIDIA's Interest Coverage Ratio changed from 2023 to 2024?}
    \item \textbf{Context:} 
    \texttt{NVIDIA Corporation and Subsidiaries...}
\end{itemize}
\textbf{Ground Truth Answer:} 
\begin{quote}
    Interest Coverage Ratio is calculated using the formula: Interest Coverage Ratio = EBIT / Interest Expenses.
        
    For 2024:\\
    EBIT {\scriptsize (Operating Income)} = \$32,972 {\scriptsize  million}\\
    Interest Expenses = \$257 {\scriptsize  million}\\
    Interest Coverage Ratio = \$32,972 / \$257 = \textbf{128.3}
    
    For 2023:\\
    EBIT {\scriptsize (Operating Income)} = \$4,224 {\scriptsize  million}\\
    Interest Expenses = \$262 {\scriptsize  million}\\
    Interest Coverage Ratio = \$4,224 / \$262 = \textbf{16.1}
\end{quote}

\textbf{Metadata:}
\begin{itemize}
    \setlength{\itemsep}{0pt}
    \item \textbf{Question Type:} Ratio Analysis
    \item \textbf{Accession Number:} 0001045810-24-000029
    \item \textbf{Page:} 50
    \item \textbf{Item:} \textit{Item 15. Exhibit and Financial Statement Schedules}
\end{itemize}

\end{tcolorbox}
}
\label{fig:example_question_no_context}
\end{figure}

~\cref{tab:summary_stats} provides summary statistics for the underlying SEC filings. In total, the questions reference 45 SEC filings from 29 different companies, fully listed in~\cref{app:accessions_list}. The supporting data spans multiple documents and may reach significant lengths, with some entries requiring tens of thousands of tokens\footnote{All token counting was done with \texttt{tiktoken.get\_encoding("cl100k\_base")}}.

\noindent\textbf{SECQUE Questions:} 
The SMEs were instructed to write questions following three main guidelines: 
I) They represent real-world questions that are interesting to a financial analyst.
II) The answers rely solely on the information provided in the reference supporting data; no external data is needed. 
III) The questions can be answered objectively, based on the provided context. The benchmark addresses four types of questions, reflecting core tasks performed by financial analysts:

(1) \textit{Risk Questions:} Financial analysts assess potential risks impacting companies based on the “Risk Factors” section of SEC filings. This task requires text analysis skills.

(2) \textit{Ratio Questions:} Analysts examine financial statements to understand a company's financial position, performance, and cash flow. This involves extracting data from tables, defining formulas, and performing calculations. 

(3) \textit{Comparison Questions:} Analysts identify trends and differences across multiple documents to evaluate a company's performance relative to peers or previous records.  

(4) \textit{Analyst Insights Questions:} Analysts synthesize multiple data points to generate conclusions and provide financial explanations. Insight questions require deep financial understanding.

~\cref{tab:dataset_breakdown} shows a breakdown of the benchmark's questions by subject.
\begin{table}[t]
\centering
\setlength{\tabcolsep}{6pt}
\caption{SECQUE breakdown by question type.}
\label{tab:dataset_breakdown}
\begin{tabular}{p{5cm}c}
\toprule
\textbf{Question Type} & \textbf{Count} \\ 
\midrule
Comparison and Trend Analysis  & 220 \\ 
Ratio Analysis                 & 188 \\ 
Risk Factors                   & 85 \\ 
Analyst Insights               & 72 \\ 
\bottomrule
\end{tabular}
\end{table}

\begin{table}[t]
\centering
\setlength{\tabcolsep}{5pt}
\caption{Token statistics by representation type.}
\label{tab:token_statistics}
\begin{tabular}{c|cccc}
\toprule
\textbf{Type} & \textbf{Mean} & \textbf{Std} & \textbf{Median} & \textbf{Max} \\ 
\midrule
HTML      & 5.4K  & 5.6K  & 3.9K   & 32.6K \\ 
Markdown  & 2.9K  & 2.9K  & 2.2K   & 16.9K \\ 
\bottomrule
\end{tabular}
\end{table}

\noindent\textbf{References to the Supporting Data:} The \textbf{context} of a question is the portion of text from an SEC filing (or multiple filings) that the SMEs have identified as relevant to answering the question. The \textbf{references to the supporting data}, indicating the pages and items to be used from each \textbf{accession number} (the unique ID of a filing), are provided in the benchmark. 

We define a \textbf{chunk} of data to be the text corresponding to a single page of the filing. If multiple chapters are covered on the same page, the chunk is divided into smaller, coherent chunks. The chunks are then concatenated to form the final context of the question, with each question requiring, on average, five chunks as context.To preserve contextual clarity when concatenating chunks, each chunk may also include a brief \textbf{header} with key information (e.g., company name, filing type, and filing date). This header slightly increases the number of tokens required to execute a question. 


\noindent\textbf{Context:} SEC filings are available for download both in XBRL and in HTML formats, and their content is composed of text and tables. We used the Markdown representation of the texts, and formatted the tables in two ways:
1) Markdown, a straightforward text-based representation that is more concise, but less expressive.
2) HTML, a structured representation using separate tags for each attribute, and styling elements removed.
~\cref{tab:token_statistics} provides key statistics about the number of tokens needed for HTML and Markdown representations, respectively.

\begin{figure}[ht]
    \centering
\includegraphics[width=0.4\textwidth]{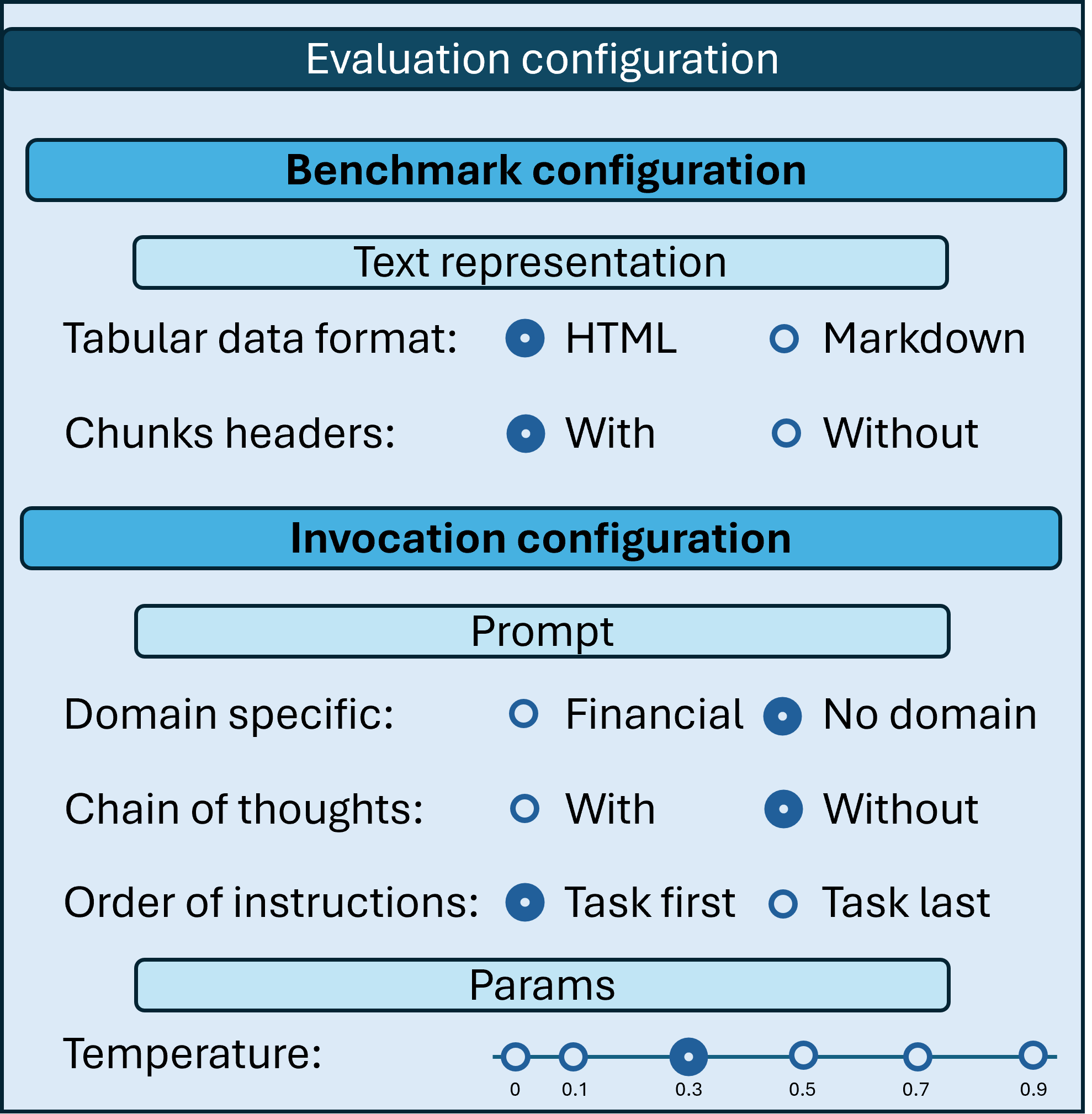}  
    \caption{Configuration for executing the SECQUE benchmark. This configuration specifies the format of the text extracted from SEC filings, along with other relevant parameters. Only one radio button can be selected within each configuration category.}
    \label{fig:configuration}
\end{figure}

Since any change in the context may impact performance on SECQUE, we provide four slightly different versions of the context for each question in the SECQUE benchmark. These versions correspond to HTML and Markdown table representations, with and without headers. ~\cref{fig:configuration} illustrates the available choices for text representation.

\section{Evaluating Judge Performance}
\label{sec:judges}

Manual evaluation of the entire benchmark is impractical, therefore, we have implemented SECQUE-judge, an automated comparison for various model outputs with the SECQUE ground truth answers (denoted as $\langle\tilde{y}, y\rangle$, respectively). 
In this section we describe our SECQUE-judge implementation and verify that it aligns well with human evaluation.

\subsection{SECQUE-judge Implementation}
For SECQUE evaluation, our primary goal is to ensure that it properly distinguishes between fully correct answers (i.e., answers acceptable for a financial analyst) and those that are partially correct or incorrect. To this end, we use \textit{Single-judge}, employing a scoring system of $\{0,1,2\}$, representing incorrect, partially correct, and correct answers, respectively. Single-judge's implementation follows the judging prompt presented in~\cite{brief2024mixing}, which similarly handles free-text comparisons categorized into three classes. We use GPT-4o~\cite{OpenAI2024} as the underlying judging model.  

Since an LLM judge can be inconsistent due to its stochastic nature, we utilize a 'panel of judges', following LLM-as-a-judge best practices outlined in ~\cite{gu2024survey}. We form our final SECQUE-judge by aggregating several Single-judge scores: for each $\langle\tilde{y}, y\rangle$ pair, we invoke Single-judge five times (using the exact same prompt and parameters). The summed score of these five individual evaluations is denoted by $ S $. SECQUE-judge maps $ S $ to a final categorical score with same $\{0,1,2\}$ scoring system using two fixed thresholds, $ U_T $ (upper threshold) and $ L_T $ (lower threshold), as defined in~\cref{eq:SECQUE_equation_params}. We aim to compute the optimal thresholds $ U_T $ and $ L_T $ for our SECQUE evaluation.

\begin{equation}\label{eq:SECQUE_equation_params}
\text{score} :=
\begin{cases}
2, & \text{if } S \geq U_T, \\
1, & \text{if } U_T > S \geq L_T, \\
0, & \text{if } S < L_T,
\end{cases}
\end{equation}

\subsection{Human Evaluation Experiment Setup}
We conducted an experiment to assess the alignment between our SECQUE-judge and expert human evaluation. First, we ran our benchmark and generated answers using GPT-4o and Llama-3.3-70B-Instruct~\cite{Dubey2024}. Due to the high cost of human evaluation, we manually selected a subset of 62 questions from all four question categories that were scored differently by several automated judges (described in ~\cref{subsec:human_evaluation_results}). Since each question was answered by two LLM models, this resulted in 124 generated answers for evaluation, 62 from GPT-4o and 62 from Llama-3.3-70B-Instruct.

Next, we presented the 124 answers to financial experts and asked them to independently compare each generated $\tilde{y}$ to its corresponding $y$ using the same $\{0,1,2\}$ scale as described earlier. This setup allows us to evaluate a lower bound on the alignment between SECQUE-judge and human evaluation.

For most questions, all human evaluators assigned the same score. In cases where the evaluation was a mix of 1 and 2, we set the final human-score to 2, as such an answer could be deemed acceptable for a financial analyst. Similarly, when scores of 0 and 1 were assigned, the final human-score was set to 0, as the answer was considered mostly incorrect. In the only four cases where evaluators disagreed entirely (with the full range of scores assigned), we set the final human-score to 1.

Since we are primarily interested in verifying that SECQUE-judge properly distinguishes fully correct answers from others, we use the following $ F_1(2) $ metric as our optimization objective:

\begin{equation}\label{eq:optimization_f1}
F_1(2) := 2 \cdot \frac{\text{precision}(2)\cdot\text{recall}(2)}{\text{precision}(2) + \text{recall}(2)},
\end{equation}
i.e., the standard multi-class $F_1$, precision, and recall scores, when 2 is the target class.

\subsection{Analyzing SECQUE-judge}\label{subsec:human_evaluation_results}
\begin{figure}[h]
   \centering
    \includegraphics[width=0.35\textwidth]{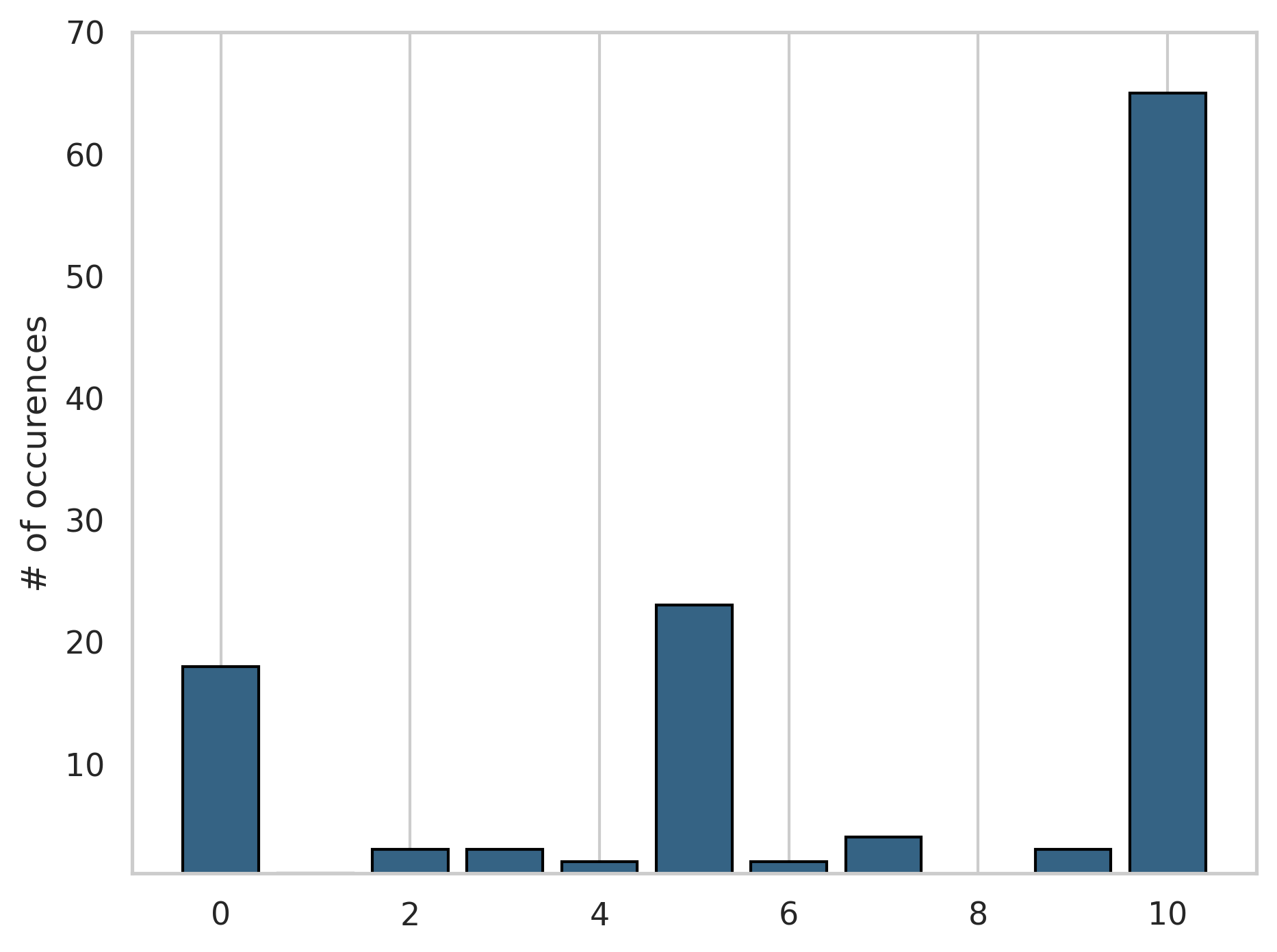}
    \caption{Histogram of $ S $, the sum of five Single-judge scores, for all 124 answers.}
    \label{fig:human_vs_judge_histogram}
\end{figure}

We begin by evaluating the stability of Single-judge scoring on the answer set. In all cases, the five Single-judge scores differed by at most 1, meaning that we did not observe both scores of 0 and 2 for the same $\langle\tilde{y}, y\rangle$ pair. In $85.5\%$ of cases, the five Single-judge scores were unanimous. ~\cref{fig:human_vs_judge_histogram} presents a histogram of $S$, the summed Single-judge scores for the 62 questions, showing that the most common sums are 0, 5, and 10, representing unanimous scores of 0, 1, and 2, respectively. 

\begin{table*}[!ht]
\centering
\small
\caption{Comparison of LLM-based judges, assessing their alignment with human judgment across multiple alignment metrics. A judge is defined both by its methodology and by the LLM used to perform the judging. The best scores for each alignment metric are indicated by underlining.}
\label{tab:Judges comparison}
\begin{tabular}{ll|cccc}
\toprule
 \multicolumn{2}{l}{Judge} &  \multicolumn{4}{c}{Alignment Metrics}\\
Methodology & Underlying Model & F1(2) & precision(2) & recall(2) & accuracy\\
\midrule
Single-judge & GPT-4o & 0.82 & 0.9 & 0.75 & 0.71\\
Majority vote & GPT-4o & 0.8 & 0.9 & 0.73 & 0.69\\
\textbf{SECQUE-judge} & \textbf{GPT-4o} & \underline{0.85} & 0.905 & 0.8 & \underline{0.75} \\
SECQUE-judge & Llama-3.3-70B-Instruct & 0.83 & 0.8 & \underline{0.86} & 0.68\\
SECQUE-judge & GPT-4o-mini & 0.62 & \underline{0.93} & 0.465 & 0.515\\
\bottomrule
\end{tabular}
\end{table*}

We then used human-scores and Single-judge summed scores $ S $ to calculate the optimal $ U_T $ and $ L_T $ (defined in~\cref{eq:SECQUE_equation_params}) maximizing our objective function $ F_1(2) $ presented in ~\cref{eq:optimization_f1}. We finalized $ U_T = 6 $ and $ L_T = 4 $ to be the threshold used in SECQUE-judge, which resulted in a maximal $ F_1(2) = 0.85 $ (the full confusion matrix is presented in~\cref{app:human evaluation experiment results}). Thus,~\cref{eq:SECQUE_evaluation_score} represents our final SECQUE-judge. It is interesting to note that $ U_T = 6 $ implies that at least one Single-judge assigned a score of 2 to the answer. Similarly, $ L_T = 4 $ implies that at least one Single-judge assigned a score of 0.  

\begin{equation}\label{eq:SECQUE_evaluation_score}
\text{score} =
\begin{cases}
2, & \text{if } S \geq 6, \\
1, & \text{if } 4 \leq S < 6, \\
0, & \text{if } S < 4.
\end{cases}
\end{equation}

Further analysis of SECQUE-judge is presented in~\cref{tab:Judges comparison}. We first observe that $\text{precision}(2) = 0.905$ and $\text{accuracy}= 0.75$. We conclude that SECQUE-judge excels in identifying fully correct answers, while its ability to distinguish between partially correct and incorrect answers is less optimal. 

SECQUE-judge also outperforms other evaluation methods in term of alignment.~\cref{tab:Judges comparison} demonstrates that employing SECQUE-judge, a panel of judges, instead of Single-judge, improves performance across all metrics by up to 4\%. Majority vote utilizes the same summed score $ S $, but results in lower alignment with human evaluation. This further implies that one Single-judge score of 2 or 0 out of five Single-judge scores is enough to award a final score of 2 and 0, respectively.

Additionally, we changed the underlying judging model, both with Llama-3.3-70B-Instruct and GPT-4o-mini~\cite{openai_gpt4o_mini}). While the first performs almost like GPT-4o, for the second we observe a significant decrease in the alignment between the judge and human evaluation. We also provide a breakdown by which model generated the answer is provided in~\cref{app:human evaluation experiment results}, to mitigate possible concerns around self-enhancement bias~\cite{Zheng2023}.

\begin{figure}[!t]
    \centering
    \includegraphics[width=0.5\textwidth]{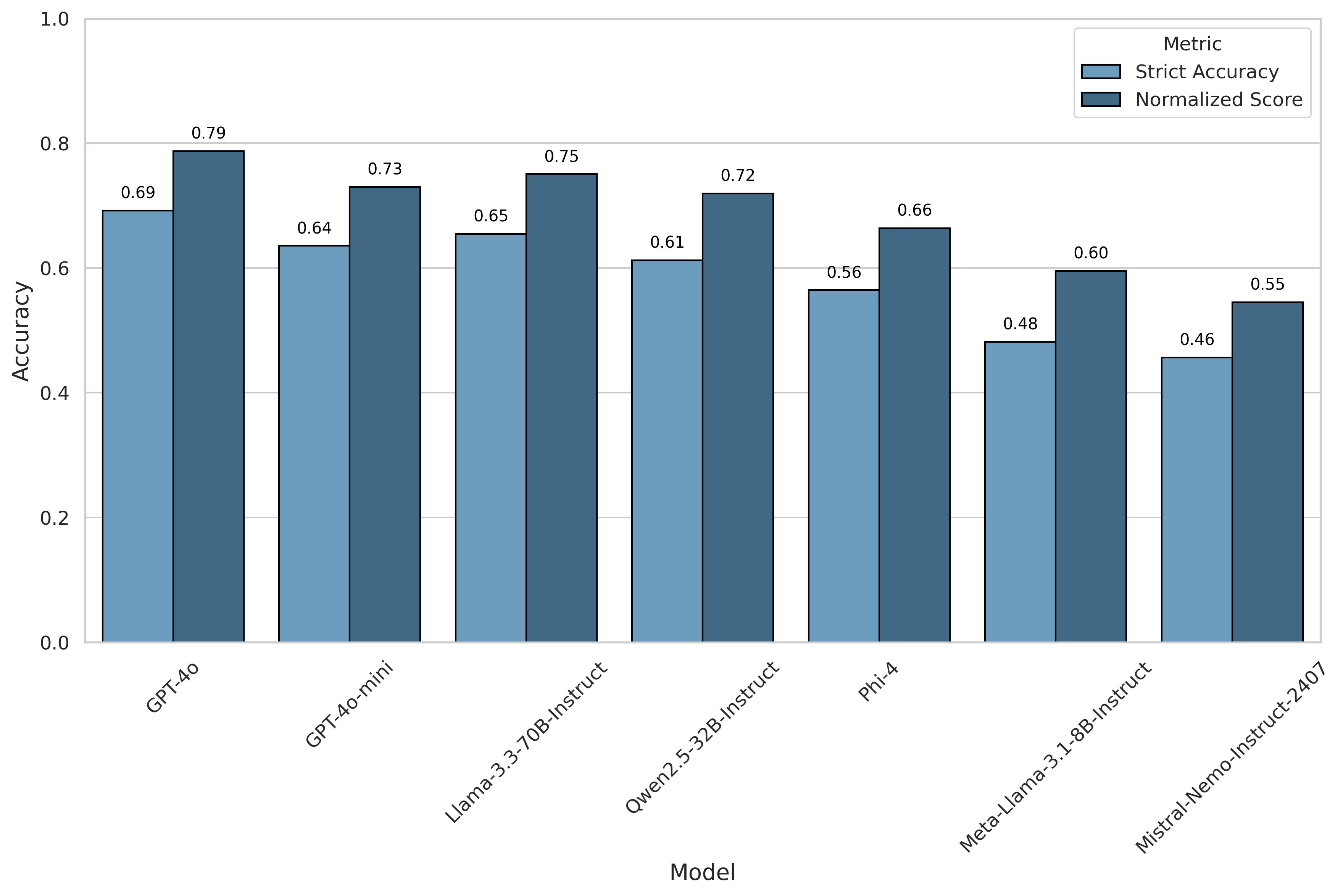}
    \caption{The performance of each model on the benchmark. Both Strict Accuracy and Normalized Accuracy are shown.}
    \label{fig:performance_baseline}
\end{figure}

\begin{figure*}[!ht]
    \centering
    \includegraphics[width=0.7\textwidth]{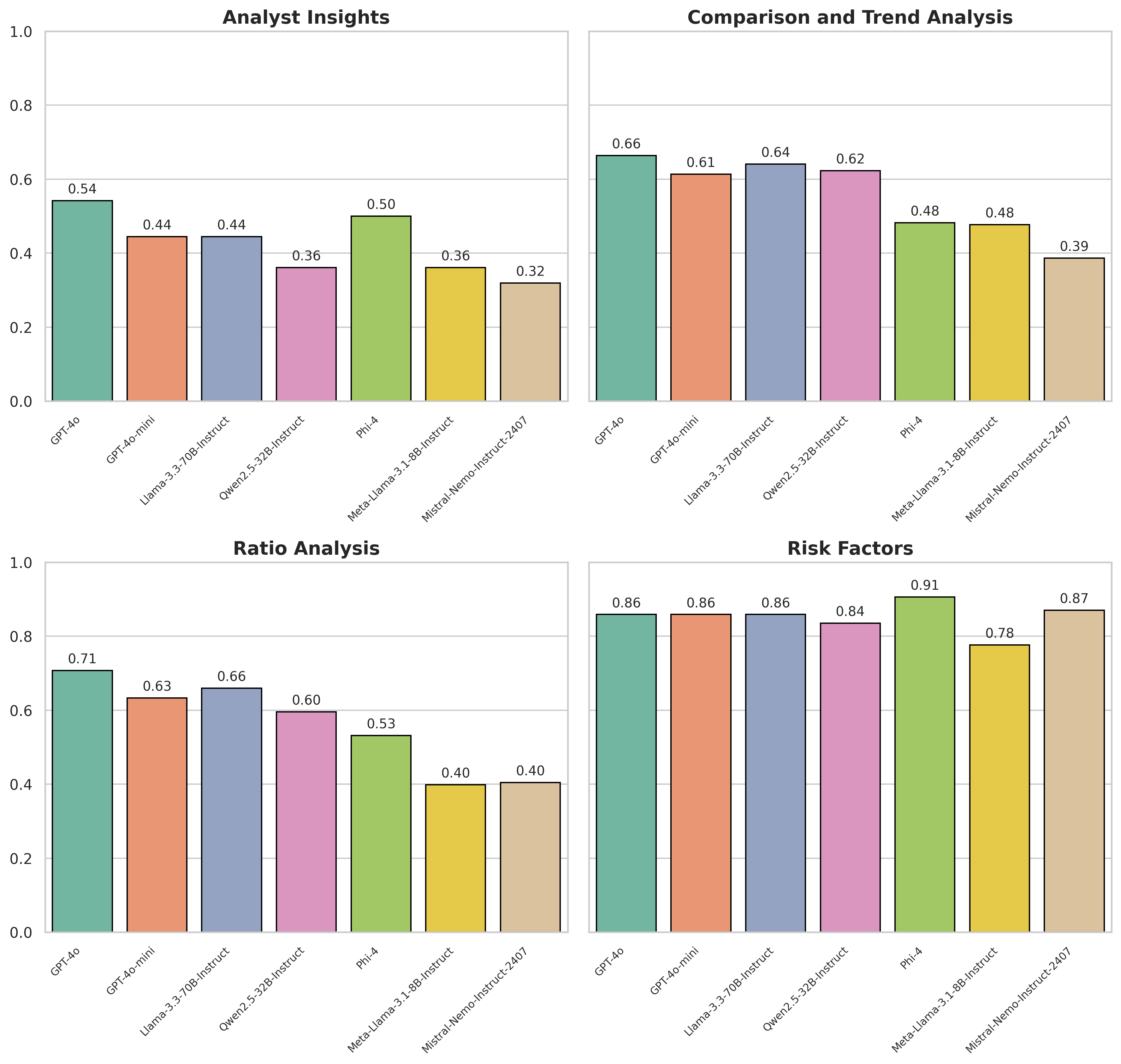}
    \caption{Model performance across different question types. Each subplot represents one question type, comparing the Strict Accuracy of all models.}
    \label{fig:performance_by_q_type}
\end{figure*}

\section{Evaluation and Results}
\label{sec:evaluation_and_results}

\begin{table*}[ht]
\centering
\scriptsize
\begin{tabular}{lcccccc}
\toprule
 & Baseline & Financial & Baseline CoT & Financial CoT & Flipped & Avg Tokens by Model \\
\midrule
GPT-4o & \textbf{\underline{0.69}}/0.79 & 0.62/0.71 & 0.67/0.76 & 0.63/0.73 & 0.68/0.78 & 319.84\\
GPT-4o-mini & \underline{0.64}/0.73 & 0.38/0.47 & 0.60/0.72 & 0.56/0.65 & 0.62/0.73 & 289.76 \\
Llama-3.3-70B-Instruct & \underline{0.65}/0.75 & 0.60/0.71 & 0.63/0.74 & 0.60/0.72 & 0.62/0.74 & 341.63\\
Qwen2.5-32B-Instruct & 0.61/0.72 & 0.49/0.58 & 0.60/0.71 & 0.55/0.67 & \underline{0.65}/0.75 & 331.34 \\
Phi-4 & 0.56/0.66 & 0.55/0.64 & \underline{0.57}/0.67 & 0.56/0.66 & \underline{0.57}/0.67 & 294.33 \\
Meta-Llama-3.1-8B-Instruct & \underline{0.48}/0.60 & 0.41/0.54 & 0.44/0.56 & 0.40/0.53 & 0.47/0.59 & 338.38\\
Mistral-Nemo-Instruct-2407 & \underline{0.46}/0.55 & 0.32/0.42 & 0.45/0.56 & 0.44/0.55 & 0.44/0.54 & 231.52\\
\midrule
Avg Tokens by Prompt & 283.04 & 151.97 & 437.38 & 334.71 & 317.57 & 304.93\\
\bottomrule
\end{tabular}
\caption{Performance metrics across prompt ablations. In each column, the left score indicates Strict Accuracy, the right Normalized Accuracy. The average number of output tokens used for each model and prompt type is included. The best score per model is underlined, and best overall is in bold}
\label{tab:all_prompt_types}
\end{table*}

\begin{figure*}[!ht]
    \centering
    \includegraphics[width=0.7\textwidth]{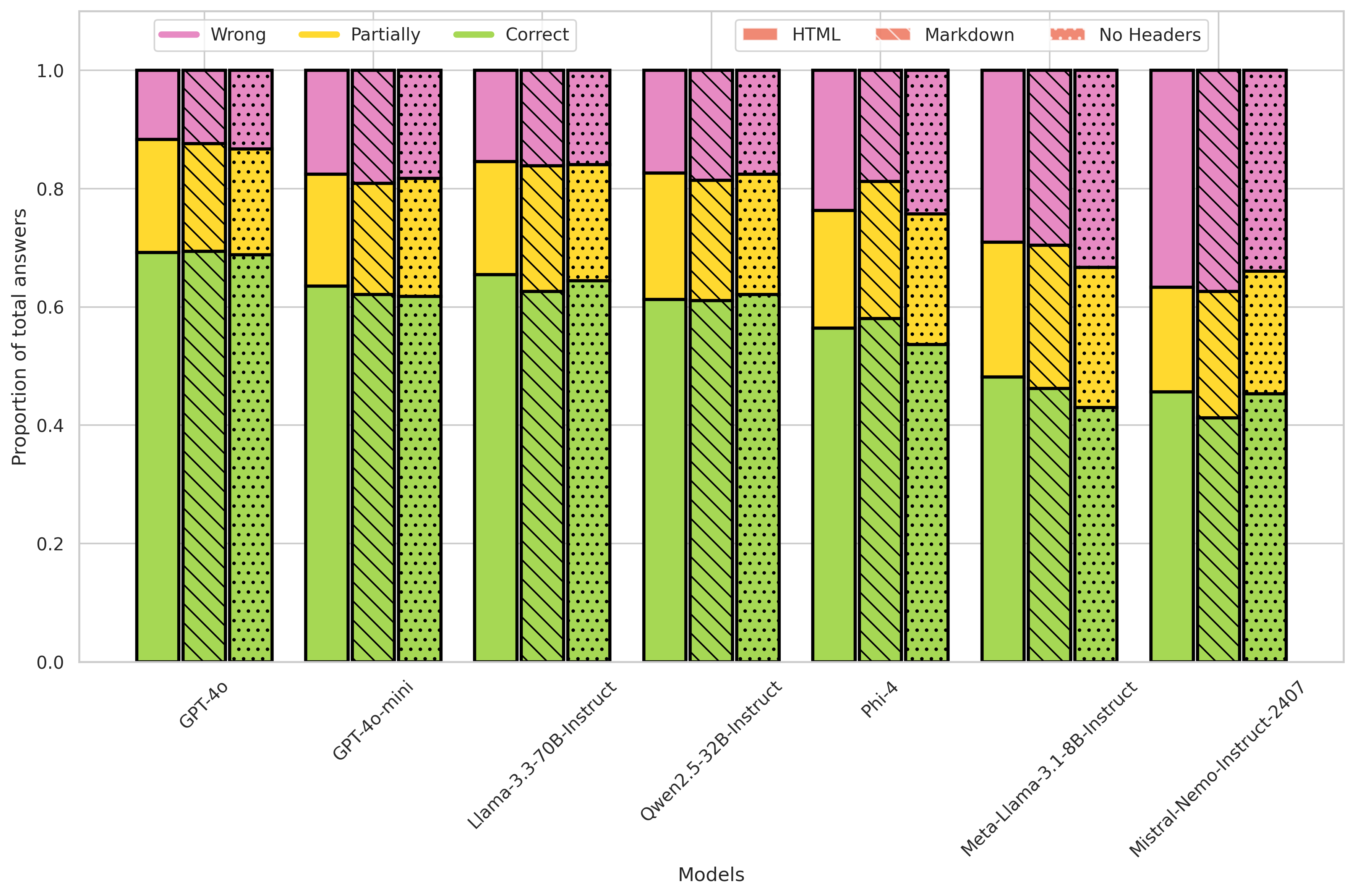}
    \caption{A comparison of all models' performance for each data representation configuration (HTML, Markdown, HTML with no headers), as well as a breakdown of scores achieved by each model. Note that the leftmost column for each model is equivalent to the baseline shown in~\cref{fig:performance_baseline}}
    \label{fig:compare_chunking_method}
\end{figure*}

\subsection{Setup}
We evaluated the performance of seven models on SECQUE, representing diverse model sizes and providers, to assess their ability to answer complex financial questions effectively. The models we chose are GPT-4o and GPT-4o-mini, Meta-Llama-3.3-70B-Instruct and Meta-Llama-3.1-8B-Instruct~\cite{Dubey2024}, Qwen2.5-32B-Instruct~\cite{qwen2.5}, Mistral-Nemo-Instruct-2407(12B)~\cite{mistral_nemo_2024}, and Phi-4(14B)~\cite{abdin2024phi}\footnote{Phi-4 has a limited context length of just 16K, resulting in lower performance, as longer questions remained unanswered.}.

All answers were scored using our SECQUE-judge. Each response was given a score according to~\cref{eq:SECQUE_evaluation_score}, which was then aggregated into two scores:

\begin{itemize}
    \item \textbf{Strict Accuracy}: $\frac{1}{2n}\sum\limits_{i} 2\mathbf{I}_{\{\text{score}=2\}}$ (2 points if score = 2 else 0).
    \item \textbf{Normalized Accuracy}: $\frac{1}{2n}\sum\limits_{i} \text{score}$ (use score directly). 
\end{itemize}

\noindent Both scores were divided by 2 to maintain a $[0,1]$ scale.

To mitigate any issues arising from the sensitivity of LLMs to input perturbations, particular attention was given to standardizing data representations and prompts.~\cref{fig:configuration} illustrates the possible configurations for an experiment using the SECQUE benchmark and identifies the 'baseline' configuration (simple prompt, \texttt{temperature=0.3}, and HTML tables with headers) that results in the highest overall performance across models. In the rest of this section we analyze the performance of the described models using the 'baseline' configuration, except for the ablation studies where we evaluate the effect of text representation, prompt and temperature configurations, both on quality and on the number of tokens produced.

\subsection{Overall Performance}

The performance of each model on the benchmark is shown in~\cref{fig:performance_baseline}. GPT-4o leads with 0.69 and 0.79 in Strict and Normalized accuracy, respectively. GPT-4o-mini and Llama-3.3-70B-Instruct have very similar performance, both slightly under GPT-4o and slightly above Qwen2.5-32B-Instruct. The smaller models perform significantly worse with Mistral-Nemo-Instruct-2407 being the furthest behind. It is interesting to note that while the absolute difference between Strict and Normalized accuracies remains similar across all models, the ratio of these accuracies is significantly higher for smaller models. This trend is more clearly illustrated in~\cref{fig:compare_chunking_method}.

\subsection{Performance by Question Type}
The various models' Strict Accuracy scores across the four SECQUE question categories are shown in~\cref{fig:performance_by_q_type}. Results highlight significant variability across categories:

\noindent\textbf{Risk Factors:} Phi-4 performed best, with almost all the other models achieving similar scores. All models achieved high scores, implying that answering such questions should be a minimum requirement for any financial model. 

\noindent\textbf{Ratio Analysis:} This category proved more challenging, with GPT-4o achieving the highest score. The results indicate both correct usage of formulas and superior mathematical reasoning abilities.

\noindent\textbf{Comparison and Trend Analysis:} The results for this category were very similar to Ratio Analysis. Smaller models exhibited difficulty reasoning over data points from long contexts, while the rest of the models had roughly equivalent performance. 

\noindent\textbf{Analyst Insights:} These questions had the lowest scores across almost all models, with GPT-4o significantly ahead, followed by Phi-4. These questions are more difficult in nature due to combining numerical reasoning and financial insights, but also involve slightly more nuanced answers, and therefore the evaluation of this category may be less reliable than the other categories.

\subsection{Ablation Study}
\noindent\textbf{Text Representation:}
The choice of text representation i.e., HTML, Markdown, and removing headers, had a small impact on overall performance. 
\cref{fig:compare_chunking_method} shows the performance of the models across two important dimensions, both comparing the representation format, and also showing a breakdown of the scores for each model. The results indicate Markdown tables were slightly harder for smaller models to interpret, indicating a trade-off between using fewer tokens and a more explicit representation format. The exception is Phi-4, gaining a boost from the token reduction due to its limited context length. The inclusion of headers is not conclusively helpful, but in most cases appears to be beneficial.

\noindent\textbf{Prompt Variations:}
Altering the prompt had the most significant impact of the various ablations. Switching from the baseline prompt to a more financial and targeted one proved to be very detrimental to performance, although better from a token usage perspective. Interestingly, while including chain-of-thought (CoT) reasoning in the baseline prompt resulted in a slight decrease in performance, incorporating CoT in the financial prompt led to a modest improvement. These findings are surprising since generally providing clearer instructions, as well as explicitly requesting the use of CoT have been shown to improve results in various reasoning tasks~\cite{wei2023chainofthoughtpromptingelicitsreasoning}. Changing the order within the prompt (context followed by question vs. question followed by context) had minimal impact, which contrasts with the findings of ~\cite{Islam2023}. This discrepancy can be attributed to our use of newer and more advanced models. All prompts can be found in~\cref{app:instruction_prompts}. 

\noindent\textbf{Temperature Settings:}
Temperature adjustments $\{0.0, 0.1, 0.3, 0.5, 0.7, 0.9\}$ were evaluated only for GPT-4o. The change in temperature had almost no impact, with less than 2\% fluctuations between values, thus we cannot conclude that the choice of temperature matters for evaluation.

\section{Related Work}
\label{sec:related_work}

Recent advances in large language models (LLMs) have spurred considerable research in domain-specific benchmarks and evaluation frameworks, particularly in finance. In this section, we briefly review work on financial benchmarks and the use of LLMs for evaluation.

\paragraph{Financial Benchmarks and Datasets}
A variety of benchmarks have been introduced to assess LLM performance on financial tasks. Comprehensive evaluation frameworks such as FinBen~\cite{Xie2024a}, PIXIU~\cite{xie2024pixiu}, and BBT-Fin~\cite{Lu2023} aggregate diverse tasks to measure general financial skills. Other datasets target specialized skills: FinEval~\cite{Zhang2023} focuses on textbook-based financial knowledge, SuperCLUE-Fin~\cite{Xu2024} decomposes real-world financial tasks into fine-grained subtasks, and FinDABench~\cite{Liu2024} emphasizes financial analysis and reasoning. In parallel, several financial QA datasets have been proposed. Early efforts include FiQA~\cite{Maia2018} for sentiment analysis and opinionated QA, while FinQA~\cite{Chen2021} and its conversational extension ConvFinQA~\cite{Chen2022} offer more realistic, multi-turn interactions. Datasets such as TAT-QA~\cite{Zhu2021} incorporate numerical reasoning over tabular and textual data from financial reports. Despite these efforts, many of the existing benchmarks do not fully capture the retrieval, analysis and reasoning challenges inherent to day-to-day financial analysis~\cite{brief2024mixing, Islam2023}, which are necessary for real-world financial work.

\paragraph{Evaluation Paradigms: LLM-as-a-Judge}
Traditional benchmark evaluation has evolved with the emergence of LLMs. Beyond standard multiple-choice or completion tasks where easy evaluation is possible, recent approaches leverage LLMs (notably GPT-4~\cite{Achiam2023}) as automated judges for assessing generation quality. For example, Li et al.~\cite{alpaca_eval} and Zheng et al.~\cite{Zheng2023} have demonstrated the effectiveness of using LLMs to score answers in open-ended question setups, while ~\cite{gu2024survey} employed majority voting from multiple judges.~\cite{gu2024survey} and others have conducted extensive studies around the alignment of LLM evaluators with human annotators, yet a single optimal setup has not been identified, prompting the need for further case-by-case optimization.

\section{Conclusions and Limitations}
\label{sec:Conclusions}
We have presented SECQUE, a comprehensive benchmark for evaluating LLMs in financial analysis tasks. Our results demonstrate that while leading models show promising capabilities in financial analysis, significant challenges remain, particularly in complex reasoning tasks and analyst insights generation. The benchmark reveals important differences in model performance across question types and highlights the critical role of configurations in evaluation results. These findings provide valuable guidance for future development of financial LLMs and evaluation frameworks.

Limitations of our work include potential biases in the LLM-based evaluation system, the need for broader coverage of financial document types. Another key limitation is that there could be more than one correct way to calculate some of the analysis questions. This is an inherent part of the domain, as there are potentially more than one way for analysts to interpret financial information. 

Future work should address these limitations by allowing for multiple correct ways to answer questions and expanding the benchmark to cover additional financial tasks and document types.

\section*{Acknowledgments}
We would like to thank Ilya Venger, Vladimir Bershtein, and Julia Korsunsky for their valuable insights and contributions throughout the process of creating the benchmark. 

\bibliography{benchmark}

\newpage
\onecolumn
\appendix
\section{Question Examples}
\label{app:question types}

\subsection*{Ratio Analysis:}
\begin{figure}[!htb]
\centering
\begin{tcolorbox}[colframe=black!40, colback=black!5!white, sharp corners=southwest, boxrule=0.6mm, toptitle=1mm, bottomtitle=1mm]
\small
\vspace{0mm}
\textbf{Input:}
\begin{itemize}
    \setlength{\itemsep}{0pt}
    \item \textbf{Question:} 
    \texttt{How has NVIDIA's Interest Coverage Ratio changed from 2023 to 2024?}
    \item \textbf{Context:} 
\end{itemize}

\begin{quote}
    NVIDIA CORP 10-K form for the fiscal year ended 2024-01-28, page 50:\\\\ 
    NVIDIA Corporation and Subsidiaries\\
    Consolidated Statements of Income\\    
    (In millions, except per share data)

\begin{tabular}{l|lr|lr|lr|}
\hline
& \multicolumn{6}{c}{Year Ended} \\
 & \multicolumn{2}{c}{Jan 28, 2024} & \multicolumn{2}{c}{Jan 29, 2023} & \multicolumn{2}{c}{Jan 30, 2022} \\
\hline
Revenue & \$ & 60,922 & \$ & 26,974 &\$ & 26,914 \\
\quad\quad Cost of revenue & & 16,621 & & 11,618 & & 9,439  \\
Gross profit & & 44,301 & &  15,356 & & 17,475 \\
Operating expenses & & & & & & \\
\quad\quad Research and development & &  8,675  & & 7,339 & & 5,268 \\
\quad\quad Sales, general and administrative & & 2,654 & & 2,440 & & 2,166\\
\quad\quad Acquisition termination cost & & - & & 1,353 & & - \\
\quad\quad\quad\quad Total operating expenses & & 11,329 &  & 11,132 & & 7,434 \\
Operating income & & 32,972 & &  4,224 & & 10,041\\
\quad\quad Interest income & & 866 & &  267 & & 29 \\
\quad\quad Interest expense & & (257) & & (262) & & (236) \\
\quad\quad Other, net & & 237 & & (48) & & 107 \\
\quad\quad\quad\quad Other income (expense), net & & 846 & & (43) & & (100)  \\
Income before income tax & & 33,818 & & 4,181 & & 9,941\\
Income tax expense (benefit) & & 4,058 & & (187) & & 189 \\
Net income & \$ & 29,760 &\$ & 4,368 & \$ & 9,752 \\
&&&&&&\\
Net income per share:  & & & & & &  \\
Basic & \$ & 12.05 & \$ & 1.76 & \$ & 3.91 \\
Diluted & \$ & 11.93 & \$ & 1.74 & \$ & 3.85 \\
&&&&&&\\
Weighted average shares used in per share computation:  & & & & & & \\
Basic & & 2,469 & & 2,487 & & 2,496 \\
Diluted & & 2,494 & & 2,507 & & 2,535 \\
\hline
\end{tabular}

See accompanying notes to the consolidated financial statements.
\end{quote}

\textbf{\\Ground Truth Answer:} 
\begin{quote}
    Interest Coverage Ratio is calculated using the formula: Interest Coverage Ratio = EBIT / Interest Expenses.
        
    For 2024:\\
    EBIT {\scriptsize (Operating Income)} = \$32,972 {\scriptsize  million}\\
    Interest Expenses = \$257 {\scriptsize  million}\\
    Interest Coverage Ratio = \$32,972 / \$257 = \textbf{128.3}
    
    For 2023:\\
    EBIT {\scriptsize (Operating Income)} = \$4,224 {\scriptsize  million}\\
    Interest Expenses = \$262 {\scriptsize  million}\\
    Interest Coverage Ratio = \$4,224 / \$262 = \textbf{16.1}
\end{quote}

\textbf{Metadata:}
\begin{itemize}
    \setlength{\itemsep}{0pt}
    \item \textbf{Question Type:} Ratio Analysis
    \item \textbf{Accession Number:} 0001045810-24-000029
    \item \textbf{Page:} 50
    \item \textbf{Item:} \textit{Item 15. Exhibit and Financial Statement Schedules}
\end{itemize}

\end{tcolorbox}
\label{fig:example_question_ratio}
\end{figure}

\newpage
\subsection*{Risk Factors:}

\begin{figure}[!htb]
\centering
\begin{tcolorbox}[colframe=black!40, colback=black!5!white, sharp corners=southwest, boxrule=0.6mm, toptitle=1mm, bottomtitle=1mm]
\small
\vspace{0mm}
\textbf{Input:}
\begin{itemize}
    \setlength{\itemsep}{0pt}
    \item \textbf{Question:} 
    \texttt{What are the potential financial and operational impacts of climate change on The Coca-Cola Company?}
    \item \textbf{Context:} 
\end{itemize}

\begin{quote}
\tiny COCA COLA CO 10-K form for the fiscal year ended 2023-12-31, page 25:\\
and oceans, as well as inefficient use of resources when packaging materials are not included in a circular economy. We and our bottling partners sell certain of our beverage products in plastic bottles and use other packaging materials that, while largely recyclable, may not be regularly recovered and recycled due to lack of collection and recycling infrastructure. If we and our bottling partners do not, or are perceived not to, act responsibly to address plastic materials recoverability and recycling concerns and associated waste management issues, our corporate image and brand reputation could be damaged, which may cause some consumers to reduce or discontinue consumption of some of our beverage products. In addition, from time to time we establish and publicly announce goals and targets to reduce the Coca-Cola system's impact on the environment by, for example, increasing our use of recycled content in our packaging materials; increasing our use of packaging materials that are made in part of plant-based renewable materials; expanding our use of reusable packaging (including refillable or returnable glass and plastic bottles, as well as dispensed and fountain delivery models where consumers use refillable containers for our beverages); participating in programs and initiatives to reclaim or recover bottles and other packaging materials that are already in the environment; and taking other actions and participating in other programs and initiatives organized or sponsored by nongovernmental organizations and other groups. If we and our bottling partners fail to achieve or improperly report on our progress toward achieving our announced environmental goals and targets, the resulting negative publicity could adversely affect consumer preference for our products. In addition, in response to environmental concerns, governmental entities in the United States and in many other jurisdictions around the world have adopted, or are considering adopting, regulations and policies designed to mandate or encourage plastic packaging waste reduction and an increase in recycling rates and\/or recycled content minimums, or, in some cases, restrict or even prohibit the use of certain plastic containers or packaging materials. These regulations and policies, whatever their scope or form, could increase the cost of our beverage products or otherwise put the Company at a competitive disadvantage. In addition, our increased focus on reducing plastic containers and other packaging materials waste has in the past and may continue to require us or our bottling partners to incur additional expenses and to increase our capital expenditures. A reduction in consumer demand for our products and\/or an increase in costs and expenditures relating to production and distribution as a result of these environmental concerns regarding plastic bottles and other packaging materials could have an adverse effect on our business and results of operations. \\
Water scarcity and poor quality could negatively impact the Coca-Cola system's costs and capacity.
Water is a main ingredient in substantially all of our products, is vital to the production of the agricultural ingredients on which our business relies and is needed in our manufacturing process. It also is critical to the prosperity of the communities we serve and the ecosystems in which we operate. Water is a limited resource in many parts of the world, facing unprecedented challenges from overexploitation, increasing demand for food and other consumer and industrial products whose manufacturing processes require water, increasing pollution and emerging awareness of potential contaminants, poor management, lack of physical or financial access to water, sociopolitical tensions due to lack of public infrastructure in certain areas of the world and the effects of climate change. As the demand for water continues to increase around the world, and as water becomes scarcer and the quality of available water deteriorates, the Coca-Cola system may incur higher costs or face capacity constraints and the possibility of reputational damage, which could adversely affect our profitability. \\
Increased demand for food products, decreased agricultural productivity and increased regulation of ingredient sourcing due diligence may negatively affect our business. \\
As part of the manufacture of our beverage products, we and our bottling partners use a number of key ingredients that are derived from agricultural commodities such as sugarcane, corn, sugar beets, citrus, coffee and tea. Increased demand for food products; decreased agricultural productivity in certain regions of the world as a result of changing weather patterns; loss of biodiversity; increased agricultural regulations, including regulation of ingredient sourcing due diligence; and other factors have in the past, and may in the future, limit the availability and\/or increase the cost of such agricultural commodities and could impact the food security of communities around the world...
Climate change and legal or regulatory responses thereto may have a long-term adverse impact on our business and results of operations.\\
There is increasing concern that a gradual increase in global average temperatures due to increased concentration of carbon dioxide and other greenhouse gases in the atmosphere is causing significant changes in weather patterns around the globe and an increase in the frequency and severity of natural disasters. Decreased agricultural productivity in certain regions of the world as a result of changing weather patterns may limit the availability or increase the cost of key agricultural commodities, such as sugarcane, corn, sugar beets, citrus, coffee and tea, which are important ingredients for our products, and could impact the food security of communities around the world. Climate change may also exacerbate extreme weather, resulting in water scarcity or flooding, and cause a further deterioration of water quality in affected regions, which could limit water availability for the Coca-Cola system's bottling operations. Increased frequency or duration of extreme weather conditions could also impair 25\\\\
COCA COLA CO 10-K form for the fiscal year ended 2023-12-31, page 26:\\
production capabilities, disrupt our supply chain or impact demand for our products. Increasing concern over climate change also may result in additional legal or regulatory requirements designed to reduce or mitigate the effects of carbon dioxide and other greenhouse gas emissions on the environment, and\/or may result in increased disclosure obligations. Increased energy or compliance costs and expenses due to increased legal or regulatory requirements may cause disruptions in, or an increase in the costs associated with, the manufacturing and distribution of our beverage products. The physical effects and transition costs of climate change and legal, regulatory or market initiatives to address climate change could have a long-term adverse impact on our business and results of operations. In addition, from time to time we establish and publicly announce goals and targets to reduce the Coca-Cola system's carbon footprint by increasing our use of recycled packaging materials, expanding our renewable energy usage, and participating in environmental and sustainability programs and initiatives organized or sponsored by nongovernmental organizations and other groups to reduce greenhouse gas emissions industrywide. If we and our bottling partners fail to achieve or improperly report on our progress toward achieving our carbon footprint reduction goals and targets, the resulting negative publicity could adversely affect consumer preference for our beverage products. \\
Adverse weather conditions could reduce the demand for our products.\\
The sales of our products are influenced to some extent by weather conditions in the markets in which we operate. Unusually cold or rainy weather during the summer months may have a temporary effect on the demand for our products and contribute to lower sales, which could have an adverse effect on our results of operations for such periods. 
\end{quote}

\textbf{\\Ground Truth Answer:} 
\begin{quote}
    Climate change poses several financial and operational risks to The Coca-Cola Company. Changes in weather patterns and increased frequency of extreme weather events can disrupt production and supply chains. For example, severe droughts or floods can impact water availability and quality, affecting manufacturing processes.
\end{quote}

\textbf{Metadata:}
\begin{itemize}
    \setlength{\itemsep}{0pt}
    \item \textbf{Question Type:} Risk Factors
    \item \textbf{Accession Number:} 0000021344-24-000009
    \item \textbf{Page:} 25, 26
    \item \textbf{Item:} \textit{ITEM 1A. RISK FACTORS}
\end{itemize}

\end{tcolorbox}
\label{fig:example_question_risk}
\end{figure}

\newpage
\subsection*{Comparison and Trend Analysis:}
\begin{figure}[!htb]
\centering
\begin{tcolorbox}[colframe=black!40, colback=black!5!white, sharp corners=southwest, boxrule=0.6mm, toptitle=1mm, bottomtitle=1mm]
\small
\vspace{0mm}
\textbf{Input:}
\begin{itemize}
    \setlength{\itemsep}{0pt}
    \item \textbf{Question:} 
    \texttt{Compare the deposit balances for Goldman Sachs and Bank of New York Mellon as of June 30, 2024.}
    \item \textbf{Context:} 
\end{itemize}

\begin{quote}
    GOLDMAN SACHS GROUP INC 10-Q form for quarterly period ended 2024-06-30, page 2:\\\\ 
    THE GOLDMAN SACHS GROUP, INC. AND SUBSIDIARIES\\
    Consolidated Balance Sheets\\
    (Unaudited)
\\\\
\tiny
\begin{tabular}{l|lr|lr|}
\hline
& \multicolumn{4}{c}{As of}\\
& & June & & December\\
\textit{\$ in millions} & & 2024 &  & 2023\\
\hline
Assets &  &  &  & \\
Cash and cash equivalents & \$ & 206,326 & \$ & 241,577 \\
Collateralized agreements: & &  &  & \\
Securities purchased under agreements to resell (includes \$198,360 and \$223,543 at fair value) & & 198,626 &  & 223,805 \\
Securities borrowed (includes \$45,819 and \$44,930 at fair value) & & 204,621 &  & 199,420\\
Customer and other receivables (includes \$23 and \$23 at fair value) & & 142,000 &  & 132,495\\
Trading assets (at fair value and includes \$117,586 and \$110,567 pledged as collateral) & & 521,981 & & 477,510 \\
Investments (includes \$86,855 and \$75,767 at fair value) & & 160,924 & & 146,839  \\
Loans (net of allowance of \$4,808 and \$5,050, and includes \$6,035 and \$6,506 at fair value) & & 184,127 &  & 183,358   \\
Other assets (includes \$243 and \$366 at fair value) &&  34,708 &  &  36,590 \\
Total assets & \$ & 1,653,313 &  \$ & 1,641,594 \\
Liabilities and shareholders' equity & &  &  & \\
Deposits (includes \$32,042 and \$29,460 at fair value) & \$ & 433,105 & \$ & 428,417 \\
Collateralized financings: & &  &  & \\
Securities sold under agreements to repurchase (at fair value) & & 238,139 &  & 249,887\\
Securities loaned (includes \$10,775 and \$8,934 at fair value) & & 63,935 &  & 60,483   \\
Other secured financings (includes \$22,868 and \$12,554 at fair value) & & 23,123 &  & 13,194 \\
Customer and other payables & & 242,986 &  & 230,728 \\
Trading liabilities (at fair value) & & 199,660 & & 200,355  \\
Unsecured short-term borrowings (includes \$49,579 and \$46,127 at fair value) & & 76,769 &  & 75,945  \\
Unsecured long-term borrowings (includes \$88,361 and \$86,410 at fair value) & & 234,632 &  & 241,877 \\
Other liabilities (includes \$142 and \$266 at fair value) & & 21,501 &  & 23,803  \\
Total liabilities & & 1,533,850 &  & 1,524,689 \\
Commitments, contingencies and guarantees &  &  &  & \\
Shareholders' equity &  & &  & \\
Preferred stock; aggregate liquidation preference of \$12,753 and \$11,203 & & 12,753 &   & 11,203  \\
Common stock; 927,414,906 and 922,895,030 shares issued, & &  & &  \\
\quad and 316,162,882 and 323,376,354 shares outstanding & & 9 &  & 9  \\
Share-based awards & &  5,058 & & 5,121 \\
Nonvoting common stock; no shares issued and outstanding &  & \textendash{} &  &\textendash{}\\
Additional paid-in capital & & 61,350 &  & 60,247 \\
Retained earnings & & 148,652 &  & 143,688  \\
Accumulated other comprehensive loss & & (2,900) &  & (2,918) \\
Stock held in treasury, at cost; 611,252,026 and 599,518,678 shares & & (105,459) &  & (100,445)  \\
Total shareholders' equity  & & 119,463  &  & 116,905  \\
Total liabilities and shareholders' equity & \$ & 1,653,313  & \$ & 1,641,594 \\
\hline
\end{tabular}
\small
\\\\See accompanying notes to the consolidated financial statements.
\end{quote}
\end{tcolorbox}
\label{fig:example_question_comparison_1}
\end{figure}

\newpage
\begin{figure}[!htb]
\centering
\begin{tcolorbox}[colframe=black!40, colback=black!5!white, sharp corners=southwest, boxrule=0.6mm, toptitle=1mm, bottomtitle=1mm]
\small
\vspace{0mm}
\begin{quote}
    Bank of New York Mellon Corp 10-Q form for quarterly period ended 2024-06-30, page 52: \\\\ 
    The Bank of New York Mellon Corporation (and its subsidiaries)\\
    Consolidated Balance Sheet (unaudited)\\\\
\tiny
\begin{tabular}{l|lr|lr|}
\hline
\textit{(dollars in millions, except per share amounts)} & & June 30, 2024 & & Dec. 31, 2023 \\
\hline
\textbf{Assets} & & & & \\
Cash and due from banks, net of allowance for credit losses of \$27 and \$18 & \$ & 5,311 & \$ & 4,922 \\
Interest-bearing deposits with the Federal Reserve and other central banks & & 116,139 & & 111,550\\
Interest-bearing deposits with banks, net of allowance for credit losses of \$1 and \$2  & &  & & \\
\quad (includes restricted of \$2,026 and \$3,420) & & 11,488 & & 12,139 \\
Federal funds sold and securities purchased under resale agreements & & 29,723 & & 28,900\\
Securities: & & & & \\
\quad Held-to-maturity, at amortized cost, net of allowance for credit losses of \$1 and \$1 & & & & \\
\quad\quad (fair value of \$41,287 and \$44,711) & & 46,429 & & 49,578\\
\quad Available-for-sale, at fair value (amortized cost of \$94,566 and \$80,678, & & & & \\
\quad\quad net of allowance for credit losses of \$5 and less than \$1) & & 90,421 & & 76,817 \\
\hline
\quad Total securities & & 136,850 & & 126,395\\
Trading assets & & 9,609 & & 10,058 \\
Loans & & 70,642 & & 66,879 \\
Allowance for credit losses & & (286) & & (303) \\
\hline
\quad Net loans & & 70,356 & & 66,576 \\
Premises and equipment & & 3,267 & & 3,163 \\
Accrued interest receivable & & 1,253 & & 1,150 \\
Goodwill & & 16,217 & & 16,261 \\
Intangible assets & & 2,826 & & 2,854 \\
Other assets, net of allowance for credit losses on accounts receivable of \$3 and \$3 & & & &  \\
\quad (includes \$1,577 and \$1,261, at fair value) & & 25,500 & & 25,909 \\
\hline
\quad\quad Total assets & \$ & 428,539 & \$ & 409,877 \\
\textbf{Liabilities} & & & & \\
Deposits: & & & & \\
\quad Noninterest-bearing deposits (principally U.S. offices) & \$ & 58,029 & \$ & 58,274 \\
\quad Interest-bearing deposits in U.S. offices & & 149,115 & & 132,616 \\
\quad Interest-bearing deposits in non-U.S. offices & & 97,167 & & 92,779 \\
\hline
\quad\quad Total deposits & & 304,311 & & 283,669\\
Federal funds purchased and securities sold under repurchase agreements & & 15,701 & & 14,507 \\
Trading liabilities & & 3,372 & & 6,226 \\
Payables to customers and broker-dealers &  & 17,569 & & 18,395 \\
Commercial paper & & 301 & & -\\
Other borrowed funds & & 280 & & 479 \\
Accrued taxes and other expenses & & 4,729 & & 5,411 \\
Other liabilities (including allowance for credit losses on lending-related commitments of \$73 and \$87, & & & & \\
\quad also includes 
\$63 and \$195, at fair value) & & 10,208 & & 9,028 \\
Long-term debt & & 30,947 & & 31,257 \\
\hline
\quad\quad Total liabilities & & 387,418 & & 368,972\\
\textbf{Temporary equity} & & & & \\
Redeemable noncontrolling interests & & 92 & & 85 \\
\textbf{Permanent equity} & && & \\
Preferred stock – par value \$0.01 per share; authorized 100,000,000 shares; issued 43,826 and 43,826 shares & & 4,343 & & 4,343 \\
Common stock – par value \$0.01 per share; authorized 3,500,000,000 shares; & & & & \\
\quad issued 1,409,173,568 and 1,402,429,447 shares & & 14 & & 14 \\
Additional paid-in capital & & 29,139 & & 28,908  \\
Retained earnings & & 40,999 & & 39,549 \\
Accumulated other comprehensive loss, net of tax & & (4,900) & & (4,893) \\
Less: Treasury stock of 671,216,069 and 643,085,355 common shares, at cost & & (28,752) & & (27,151) \\
\hline
\quad\quad Total The Bank of New York Mellon Corporation shareholders' equity & & 40,843 & & 40,770 \\
Nonredeemable noncontrolling interests of consolidated investment management funds & & 186 & & 50 \\
\hline
\quad\quad Total permanent equity & & 41,029 & & 40,820\\
\hline
\quad\quad Total liabilities, temporary equity and permanent equity & \$ & 428,539 & \$ & 409,877 \\
\hline
\end{tabular}
\small
\\\\See accompanying unaudited Notes to Consolidated Financial Statements
\end{quote}
\textbf{\\Ground Truth Answer:} 
\begin{quote}
    As of June 30, 2024, Goldman Sachs' deposits were \$433,105 million, up from \$428,417 million as of December 31, 2023, marking a 1.1\% increase. Bank of New York Mellon's total deposits were \$304,311 million as of June 30, 2024, up from \$283,669 million as of December 31, 2023, marking a 7.3\% increase.
\end{quote}

\textbf{Metadata:}
\begin{itemize}
    \setlength{\itemsep}{0pt}
    \item \textbf{Question Type:} Comparison and Trend Analysis
    \item \textbf{Accession Number:} 0000886982-24-000022;\quad 0001390777-24-000105
    \item \textbf{Page:} 2;\quad 52
    \item \textbf{Item:} \textit{Item 1. Financial Statements (Unaudited)};\quad\textit{Item 1. Financial Statements:}
\end{itemize}

\end{tcolorbox}
\label{fig:example_question_comparison_2}
\end{figure}

\newpage
\subsection*{Analyst Insights:}
\begin{figure}[!htb]
\centering
\begin{tcolorbox}[colframe=black!40, colback=black!5!white, sharp corners=southwest, boxrule=0.6mm, toptitle=1mm, bottomtitle=1mm]
\small
\vspace{0mm}
\textbf{Input:}
\begin{itemize}
    \setlength{\itemsep}{0pt}
    \item \textbf{Question:} 
    \texttt{How does DFS Debt-to-Equity Ratio for 2023 reflect on the company's financial stability?}
    \item \textbf{Context:} 
\end{itemize}

\begin{quote}
\scriptsize 
    Discover Financial Services 10-K form for the fiscal year ended 2023-12-31, page 85:\\\\ 
\tiny
    DISCOVER FINANCIAL SERVICES\\
    Consolidated Statements of Financial Condition\\
    (dollars in millions, except for share amounts)\\
\tiny
\begin{tabular}{l|lr|lr|}
& \multicolumn{4}{c}{\textbf{December 31}} \\
& \multicolumn{2}{c}{\textbf{2023}} & \multicolumn{2}{c}{\textbf{2022}} \\
\hline
\textbf{Assets} & & & & \\
Cash and cash equivalents & \$ & 11,685 & \$ & 8,856 \\
Restricted cash & & 43 & & 41 \\
Investment securities (includes available-for-sale securities of \$13,402 and \$11,987 & & & & \\
\quad reported at fair value with associated amortized cost of \$13,451 and \$12,167 & & & &  \\
\quad at December 31, 2023 and 2022, respectively) & & 13,655 & & 12,208 \\
Loan receivables & & & & \\
\quad\quad Loan receivables & & 128,409 & & 112,120 \\
\quad\quad Allowance for credit losses & & (9,283) & & (7,374) \\
Net loan receivables & & 119,126 & & 104,746 \\
Premises and equipment, net & & 1,091 & & 1,003 \\
Goodwill & & 255 & & 255 \\
Other assets & & 5,667 & & 4,597 \\
Total assets & & 151,522 & & 131,706 \\
\textbf{Liabilities and Stockholders' Equity} & & & & \\
\textit{Liabilities} & & & & \\
Deposits & & & & \\
\quad\quad Interest-bearing deposit accounts & & 107,493 & & 90,151 \\
\quad\quad Non-interest-bearing deposit accounts & & 1,438 & & 1,485 \\
Total deposits & & 108,931 & & 91,636 \\
Short-term borrowings & & 750 & & - \\
Long-term borrowings & & 20,581 & & 20,108 \\
Accrued expenses and other liabilities & & 6,432 & & 5,618 \\
Total liabilities & & 136,694 & & 117,362 \\
Commitments, contingencies and guarantees \tiny{(Notes 15, 18 and 19)} & & & & \\
\textbf{Stockholders' Equity} & & & & \\
Common stock, par value \$0.01 per share; 2,000,000,000 shares authorized;  & & & & \\
\quad 570,837,720 and 569,689,007 shares issued at December 31, 2023 and 2022, respectively & &6 & & 6\\
Preferred stock, par value \$0.01 per share; 200,000,000 shares authorized; & & & & \\
\quad 10,700 shares issued and outstanding at December 31, 2023 and 2022, respectively & & 1,056 & & 1,056 \\
Additional paid-in capital & & 4,553 & & 4,468 \\
Retained earnings & & 30,448 & & 28,207 \\
Accumulated other comprehensive loss & & (225) & & (339) \\
Treasury stock, at cost; 320,734,860 and 302,305,216 shares & & & & \\
\quad at December 31, 2023 and 2022, respectively & & (21,010) & & (19,054) \\
Total stockholders' equity & & 14,828 & & 14,344 \\
Total liabilities and stockholders' equity & & 151,522 & & 131,706 \\
\hline
\end{tabular}
\scriptsize 
\\\\
The table below presents the carrying amounts of certain assets and liabilities of Discover Financial Services' consolidated variable interest entities (\"VIEs\"), which are included in the consolidated statements of financial condition above. The assets in the table below include those assets that can only be used to settle obligations of the consolidated VIEs. The liabilities in the table below include third-party liabilities of consolidated VIEs only and exclude intercompany balances that eliminate in consolidation. The liabilities also exclude amounts for which creditors have recourse to the general credit of Discover Financial Services.
\\\\
\tiny
\begin{tabular}{l|lr|lr|}
\hline
& \multicolumn{4}{c}{December 31}\\
& \multicolumn{2}{c}{2023} & \multicolumn{2}{c}{2022}\\
\hline
\textbf{Assets} &  &  &  & \\
Restricted cash & \$ & 43 & \$ & 41 \\
Loan receivables & \$ & 30,590 & \$ & 25,937 \\
Allowance for credit losses allocated to securitized loan receivables & \$ & (1,347) & \$ & (1,152) \\
Other assets & \$ & 3 & \$ & 3 \\
\textbf{Liabilities} &  &  &  & \\
Short- and long-term borrowings & \$ & 11,743 & \$ & 10,259 \\
Accrued expenses and other liabilities & \$ & 19 & \$ & 14 \\
\hline
\end{tabular}
\\\\
See Notes to the Consolidated Financial Statements
\end{quote}
\textbf{\\Ground Truth Answer:} 
\begin{quote}
\scriptsize
    Increase in Leverage: The ratio increased from 8.2 in 2022 to 9.2 in 2023, indicating higher reliance on debt relative to equity. \\
    Financial Risk: The higher ratio suggests greater financial risk due to increased debt obligations.\\
    Impact on Stability: Greater leverage could affect financial stability, especially in adverse economic conditions or with rising interest rates.
\end{quote}

\textbf{Metadata:}
\begin{itemize}
    \setlength{\itemsep}{0pt}
    \item \textbf{Question Type:} Analyst Insights
    \item \textbf{Accession Number:} 0001393612-24-000010
    \item \textbf{Page:} 85
    \item \textbf{Item:} \textit{Item 8.    Financial Statements and Supplementary Data}
\end{itemize}
\end{tcolorbox}
\label{fig:example_question_analysis}
\end{figure}

\newpage
\section{Instruction Prompts}
\label{app:instruction_prompts}
The various prompts from~\cref{tab:all_prompt_types} are included here.

\begin{figure*}[!htb]
\centering
\begin{tcolorbox}
\fbox{Baseline Prompt}\par \vspace{1mm}
You are given a financial question and a financial document. Your task is to answer the question based on the document.

\vspace{5mm}
\textbf{Input:}
\begin{itemize}
    \item \textbf{Document:} 
    \texttt{\{document\}}
    \item \textbf{Question:} 
    \texttt{\{question\}}
\end{itemize}

\textbf{Output:}
\begin{itemize}
    \item \textit{A response answering the question based on the provided document.}
\end{itemize}

\end{tcolorbox}
\label{fig:baseline_prompt}
\end{figure*}

\begin{figure*}[!htb]
\centering
\begin{tcolorbox}
\fbox{Financial Prompt}\par \vspace{1mm}
You are given a financial text extracted from 10-K or 10-Q files and a question written by domain experts. Your task is to answer the question based only on the provided context. Do not use any additional context. Your answer should be concise and accurate. In case you are unable to answer the question, you should state that you can't answer the question. Do not guess and do not suggest your own solutions.

\vspace{5mm}
\textbf{Input:}
\begin{itemize}
    \item \textbf{Document:} 
    \texttt{\{document\}}
    \item \textbf{Question:} 
    \texttt{\{question\}}
\end{itemize}

\textbf{Output:}
\begin{itemize}
    \item \textit{A response answering the question based on the provided document.}
\end{itemize}

\end{tcolorbox}
\label{fig:financial_prompt}
\end{figure*}

\begin{figure*}[!htb]
\centering
\begin{tcolorbox}
\fbox{Baseline Prompt with CoT}\par \vspace{1mm}
You are given a financial question and a financial document. Your task is to answer the question based on the document. Think step-by-step, and describe your reasoning process clearly before providing the final answer. You must provide the correct answer in a clear manner. Begin by describing your detailed reasoning process in a step-by-step manner, and then provide the final answer.

\vspace{5mm}
\textbf{Input:}
\begin{itemize}
    \item \textbf{Document:} 
    \texttt{\{document\}}
    \item \textbf{Question:} 
    \texttt{\{question\}}
\end{itemize}

\textbf{Output:}
\begin{itemize}
    \item \textit{A response answering the question based on the provided document, including a step-by-step reasoning process.}
\end{itemize}

\end{tcolorbox}
\label{fig:baseline_prompt_cot}
\end{figure*}

\begin{figure*}[!htb]
\centering
\begin{tcolorbox}
\fbox{Financial Prompt with CoT}\par \vspace{1mm}
You are given a financial text extracted from 10-K or 10-Q files and a question written by domain experts. Your task is to answer the question based only on the provided context. Do not use any additional context. Your answer should be concise and accurate. In case you are unable to answer the question, you should state that you can't answer the question. Do not guess and do not suggest your own solutions. Think step-by-step, and describe your reasoning process clearly before providing the final answer. You must provide the correct answer in a clear manner. Begin by describing your detailed reasoning process in a step-by-step manner, and then provide the final answer.

\vspace{5mm}
\textbf{Input:}
\begin{itemize}
    \item \textbf{Document:} 
    \texttt{\{document\}}
    \item \textbf{Question:} 
    \texttt{\{question\}}
\end{itemize}

\textbf{Output:}
\begin{itemize}
    \item \textit{A response answering the question based on the provided document, including a step-by-step reasoning process.}
\end{itemize}

\end{tcolorbox}
\label{fig:financial_prompt_cot}
\end{figure*}

\clearpage
\section{Human Evaluation Experiment results}
\label{app:human evaluation experiment results}
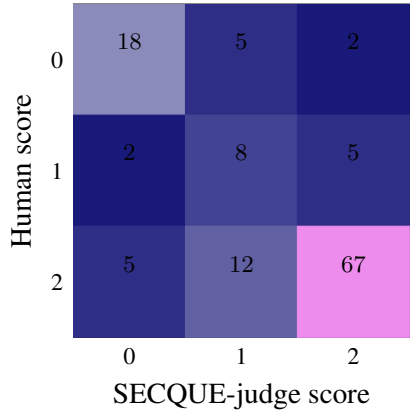
\begin{figure}[!h]
    \centering
    \begin{tikzpicture}
        \begin{axis}[
            colormap/violet,
            width=6cm,
            height=6cm,
            xtick={0,1,2},
            ytick={2,1,0},
            xticklabels={0,1,2},
            yticklabels={0,1,2},
            xlabel={SECQUE-judge score},
            ylabel={Human score},
            enlargelimits=false,
            tick label style={font=\footnotesize},
            nodes near coords,
            every node near coord/.append style={font=\footnotesize, black},
            point meta=explicit
        ]
        \addplot [
            matrix plot*,
            mesh/cols=3
        ] table [meta=C] {
            X Y C
            0 2 18
            1 2 5
            2 2 2
            0 1 2
            1 1 8
            2 1 5
            0 0 5
            1 0 12
            2 0 67
        };
        \end{axis}
    \end{tikzpicture}
    \caption{Confusion matrix heatmap comparing human scores to SECQUE-judge scores.}
    \label{fig:confusion_matrix_heatmap}
\end{figure}

We provide additional details about our judge alignment experiment.~\cref{fig:confusion_matrix_heatmap} displays the detailed confusion matrix of our LLM judge relative to human scores, and~\cref{tab:SECQUE-Judge stability} show the stability of the LLM judge across two different models' outputs.


\begin{table}[!h]
    \centering
    \setlength{\tabcolsep}{5pt}
    \caption{Stability test for SECQUE-Judge for the 62 outputs from each model. \textit{Both} is the average for all 124 (as shown in~\cref{tab:Judges comparison}.)}
    \label{tab:SECQUE-Judge stability}
    \begin{tabular}{lc|cccc}
    \toprule
     Data source & \#Answers & \multicolumn{4}{c}{Alignment Metrics}\\
    & &  F1(2) & precision(2) & recall(2) & accuracy\\
    \midrule
    Both & 124 & 0.85 & 0.905 & 0.8 & 0.75 \\
    GPT-4o & 62 & 0.86  & 0.895 & 0.83 & 0.76 \\
    Llama-3.3-70B & 62 & 0.835 & 0.915 & 0.77 & 0.74 \\
    \bottomrule
    \end{tabular}
\end{table}

\newpage
\section{Full List of Accessions}
\label{app:accessions_list}
\cref{tab:accession_numbers} lists the exact filings used in SECQUE.

\begin{table*}[!hbt]
\centering
\small
\caption{Accession Numbers and Filing Periods}
\label{tab:accession_numbers}
\begin{tabular}{|p{3.8cm}|p{8cm}|p{0.8cm}|p{1.8cm}|}
\hline
\textbf{Accession Number} & \textbf{Company Name} & \textbf{From} & \textbf{Filing Date} \\ \hline
0000004962-24-000052 & AMERICAN EXPRESS CO & 10-Q & 2024-07-19  \\ \hline
0000004962-24-000013 & AMERICAN EXPRESS CO & 10-K & 2024-02-09  \\ \hline
0000732717-24-000009 & AT\&T INC. & 10-K & 2024-02-23  \\ \hline
0000320193-24-000081 & Apple Inc. & 10-Q & 2024-08-02  \\ \hline
0000320193-24-000069 & Apple Inc. & 10-Q & 2024-05-03  \\ \hline
0000320193-23-000106 & Apple Inc. & 10-K & 2023-11-03  \\ \hline
0000320193-22-000108 & Apple Inc. & 10-K & 2022-10-28  \\ \hline
0000070858-24-000208 & BANK OF AMERICA CORP /DE/ & 10-Q & 2024-07-30  \\ \hline
0000070858-24-000156 & BANK OF AMERICA CORP /DE/ & 10-Q & 2024-04-30  \\ \hline
0001390777-24-000105 & Bank of New York Mellon Corp & 10-Q & 2024-08-02  \\ \hline
0000093410-24-000040 & CHEVRON CORP & 10-Q & 2024-08-07  \\ \hline
0000811156-24-000084 & CMS ENERGY CORP & 10-Q & 2024-04-25  \\ \hline
0000021344-24-000044 & COCA COLA CO & 10-Q & 2024-07-29  \\ \hline
0000021344-24-000009 & COCA COLA CO & 10-K & 2024-02-20  \\ \hline
0001393612-24-000047 & Discover Financial Services & 10-Q & 2024-07-31  \\ \hline
0001393612-24-000010 & Discover Financial Services & 10-K & 2024-02-23  \\ \hline
0000034088-24-000050 & EXXON MOBIL CORP & 10-Q & 2024-08-05  \\ \hline
0001262039-24-000037 & Fortinet, Inc. & 10-Q & 2024-08-08  \\ \hline
0001262039-24-000014 & Fortinet, Inc. & 10-K & 2024-02-26  \\ \hline
0001562762-24-000034 & Frontier Communications Parent, Inc. & 10-K & 2024-02-23  \\ \hline
0001193125-24-168943 & GENERAL MILLS INC & 10-K & 2024-06-26  \\ \hline
0001193125-23-177500 & GENERAL MILLS INC & 10-K & 2023-06-28  \\ \hline
0000886982-24-000022 & GOLDMAN SACHS GROUP INC & 10-Q & 2024-08-02  \\ \hline
0000886982-24-000016 & GOLDMAN SACHS GROUP INC & 10-Q & 2024-05-03  \\ \hline
0000886982-23-000011 & GOLDMAN SACHS GROUP INC & 10-Q & 2023-11-03  \\ \hline
0000045012-24-000007 & HALLIBURTON CO & 10-K & 2024-02-06  \\ \hline
0000773840-24-000051 & HONEYWELL INTERNATIONAL INC & 10-Q & 2024-04-25  \\ \hline
0000051143-24-000012 & INTERNATIONAL BUSINESS MACHINES CORP & 10-K & 2024-02-26  \\ \hline
0000091419-24-000054 & J M SMUCKER Co & 10-K & 2024-06-18  \\ \hline
0000091419-22-000049 & J M SMUCKER Co & 10-K & 2022-06-16  \\ \hline
0000200406-24-000013 & JOHNSON \& JOHNSON & 10-K & 2024-02-16  \\ \hline
0000019617-24-000453 & JPMORGAN CHASE \& CO & 10-Q & 2024-08-02  \\ \hline
0000019617-24-000326 & JPMORGAN CHASE \& CO & 10-Q & 2024-05-01  \\ \hline
0000019617-24-000225 & JPMORGAN CHASE \& CO & 10-K & 2024-02-16  \\ \hline
0000753308-24-000008 & NEXTERA ENERGY INC & 10-K & 2024-02-16  \\ \hline
0000320187-18-000142 & NIKE INC & 10-K & 2018-07-25  \\ \hline
0001045810-24-000029 & NVIDIA CORP & 10-K & 2024-02-21  \\ \hline
0000078003-24-000166 & PFIZER INC & 10-Q & 2024-08-05  \\ \hline
0000080424-24-000083 & PROCTER \& GAMBLE Co & 10-K & 2024-08-05  \\ \hline
0000080424-23-000073 & PROCTER \& GAMBLE Co & 10-K & 2023-08-04  \\ \hline
0001560327-24-000021 & Rapid7, Inc. & 10-K & 2024-02-26  \\ \hline
0001558370-24-001532 & SIMON PROPERTY GROUP INC /DE/ & 10-K & 2024-02-22  \\ \hline
0001628280-24-002390 & Tesla, Inc. & 10-K & 2024-01-29  \\ \hline
0000950170-22-000796 & Tesla, Inc. & 10-K & 2022-02-07  \\ \hline
0000899689-24-000005 & VORNADO REALTY TRUST & 10-K & 2024-02-12  \\ \hline
\end{tabular}
\end{table*}

\end{document}